# Mixture-of-Parents Maximum Entropy Markov Models

**David S. Rosenberg**
Department of Statistics
University of California, Berkeley
drosen@stat.berkeley.edu

**Dan Klein**
Computer Science Division
University of California, Berkeley
klein@cs.berkeley.edu

**Ben Taskar**
Computer and Information Science
University of Pennsylvania
taskar@cis.upenn.edu

## Abstract

We present the mixture-of-parents maximum entropy Markov model (MoP-MEMM), a class of directed graphical models extending MEMMs. The MoP-MEMM allows tractable incorporation of long-range dependencies between nodes by restricting the conditional distribution of each node to be a mixture of distributions given the parents. We show how to efficiently compute the exact marginal posterior node distributions, regardless of the range of the dependencies. This enables us to model non-sequential correlations present within text documents, as well as between interconnected documents, such as hyperlinked web pages. We apply the MoP-MEMM to a named entity recognition task and a web page classification task. In each, our model shows significant improvement over the basic MEMM, and is competitive with other long-range sequence models that use approximate inference.

## 1 Introduction

Two very popular and effective techniques for sequence labeling tasks, such as part-of-speech tagging, are maximum entropy Markov models (MEMMs), introduced by McCallum et al. [2000], and linear-chain conditional random fields (CRFs), introduced by Lafferty et al. [2001]. Neither of these models directly model relationships between nonadjacent labels. Higher order Markov models relax this local conditional independence assumption, but the complexity of inference grows exponentially with the increasing range of direct dependencies. In many situations, models could benefit by allowing information to pass directly between two labels that are far apart. For example, in named entity recognition (NER) tasks, a typical goal is to identify groups of consecutive words as being one of the following entity types: location, person, company, and other. It often happens that the type of an entity is clear in one context, but difficult to determine in another context. In a Markov model of fixed order, there is no direct way to share information between the two occurrences of the same entity. However, with long-distance interactions, we can enforce or encourage repeated words and word groups to receive the same entity labels.

Long-range dependencies arise not only within contiguous text, but also between interconnected documents. Consider the task of giving a topic label to each document in a collection, where the documents have a natural connectivity structure. For example, in a collection of scientific articles, it's natural to consider there to be a "connection" between two articles if one article cites the other. Similarly, for a collection of web pages, a hyperlink from one web page to another is a natural indicator of connection. Since documents often connect to other documents about similar topics, it's potentially helpful to use this connectivity information in making topic label predictions. Indeed, this structure has been used to aid classification in several non-probabilistic, procedural systems [Neville and Jensen, 2000, Slattery and Mitchell, 2000], as well as in probabilistic models [Getoor et al., 2001, Taskar et al., 2002, Bunescu and Mooney, 2004].

Although a strong case can be made for the benefits of long-range models, performing inference (i.e. carrying out the labeling procedure) is intractable in most graphical models with long-range interactions. One general approach to this challenge is to replace exact inference with approximate inference algorithms. Two previous approaches to using long-distance dependencies in linguistic tasks are loopy belief propagation [Taskar et al., 2002, Sutton and McCallum, 2004, Bunescu and Mooney, 2004] and Gibbs sampling [Finkel et al., 2005], each a form of approximate inference.

In this paper, we present the *mixture-of-parents MEMM*, a graphical model incorporating long-range interactions, for which we can efficiently compute marginal node posteriors without approximation or sampling. As a graphical model, the mixture-of-parents MEMM is a MEMM with additional "skip edges" that connect nonadjacent nodes. The skip edges are directed edges pointing from earlier labels to later labels. At this level, the model is similar to the skip-chain CRF of Sutton and McCallum [2004], which they describe as essentially a linear-chain CRF with additional long-distance edges. However, while the skip-chain CRF precludes exact inference, we make additional model assumptions to keep exact inference tractable.

In both mixture-of-parents MEMMs and skip-chain CRFs, the features on skip edges can be based on both the label and the input environment of each node in the edge. These long-distance features allow a highly informative environment at one node to influence the label chosen for the other node. In the NER task, for example, one might connect all pairs of identical words by edges. This would allow the context-sharing effect described above. In linked-document data, a number of interesting models are possible. One simple model is to have a long-distance feature connecting each document to all the other documents that it cites [Chakrabarti et al., 1998, Taskar et al., 2001]. These links allow the model to account for the strong topic correlation along bibliographic links.

The rest of the paper is organized as follows. We begin in Section 2 by reviewing maximum entropy Markov models. Then we introduce the mixture-of-parents extension, and show how to perform efficient inference in the model. In Section 3, we describe the estimation procedure. In Sections 4 and 5, we provide experimental validation on two tasks, demonstrating significant improvements in accuracy. We conclude with a discussion of our results in Section 6.

## 2 Models

We begin by describing maximum entropy Markov models (MEMMs), introduced by McCallum et al. [2000]. A MEMM represents the conditional distribution of a chain of labels given the input sequence.

### 2.1 MEMMs

Let us denote the input sequence as $\mathbf{x} = (x_1, x_2, \ldots, x_n)$ and the label sequence as $\mathbf{y} = (y_1, y_2, \ldots, y_n)$, where each label $y_i$ takes on values in some discrete set $\mathcal{Y}$. A first-order MEMM assumes

$$p(y_k \mid y_1, \ldots, y_{k-1}, \mathbf{x}) = p(y_k \mid y_{k-1}, \mathbf{x}).$$

Inference in these models, that is, computing the posterior marginals $p(y_1 \mid x), \ldots, p(y_n \mid x)$, or the posterior mode

$$\arg\max_{y_1, \ldots, y_n} p(y_1, \ldots, y_n \mid \mathbf{x}),$$

requires $O(n|\mathcal{Y}|^2)$ time. An $m^{th}$-order model assumes

$$p(y_k \mid y_1, \ldots, y_{k-1}, \mathbf{x}) = p(y_k \mid y_{k-m}, \ldots, y_{k-1}, \mathbf{x}),$$

and requires $O(n|\mathcal{Y}|^{m+1})$ time for inference. For simplicity, we focus on first-order models. In a MEMM, the conditional distributions $p(y_k \mid y_{k-1}, \mathbf{x})$ are taken to be log-linear, or "maximum entropy" in form:

$$p_{\lambda,\mu}(y_k \mid y_{k-1}, \mathbf{x}) = \frac{1}{Z_{y_{k-1},\mathbf{x}}} \exp\left(\sum_s \lambda_s f_s(y_{k-1}, y_k, \mathbf{x}) + \sum_t \mu_t g_t(y_k, \mathbf{x})\right)$$

where $Z_{y_{i-1},\mathbf{x}}$ is a normalization function ensuring that $\sum_{y_i \in \mathcal{Y}} p_{\lambda,\mu}(y_i \mid y_{i-1}, \mathbf{x}) = 1$. In models for named entity recognition, the features $f_s$ and $g_t$ track the attributes of the local context around the label, such as the current word, previous words, capitalization, punctuation, etc.

### 2.2 Skip-chain models

In the MEMM, when we condition on the input sequence $\mathbf{x}$, the label variables $y_1, \ldots, y_n$ form a Markov chain. The conditional independence structure of this model is given by a directed graph, with edges connecting adjacent labels. We now consider a much more general model, in which we allow additional "skip" edges to connect nonadjacent labels. We call this model a directed skip-chain model. The graphical structure for this model is shown in Figure 1(a), with the long-range skip edges shown dashed. With respect to the graph structure, the parents of a label $y_k$ comprise the label $y_{k-1}$ immediately preceding it, as well as any earlier labels connected to $y_k$ via a skip edge.

For each label $y_k$, we denote the indices of the parents of $y_k$ by $\pi_k \subseteq \{1, \ldots, k-1\}$, and we denote the set of parent labels by $\mathbf{y}_{\pi_k} = \{y_j : j \in \pi_k\}$. We define the conditional distribution of $y_k$ as follows:

$$p(y_k \mid y_1, \ldots, y_{k-1}, \mathbf{x}) = p(y_k \mid \mathbf{y}_{\pi_k}, \mathbf{x}). \quad (1)$$

Since we are conditioning on the input $\mathbf{x}$, the graphical structure itself is allowed to depend on the input. This allows us, for instance, to introduce skip edges connecting the labels of identical words. An undirected version of this model, called the skip-chain conditional random field, has been presented in [Sutton and McCallum, 2004]. Figure 1(b) shows the graphical structure of the skip-chain CRF.

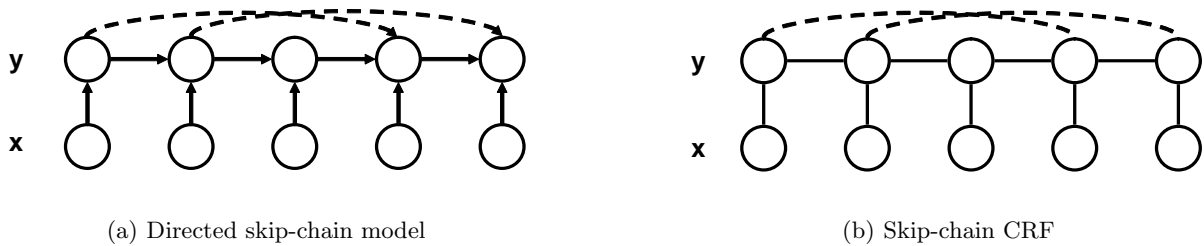

(a) Directed skip-chain model    (b) Skip-chain CRF

Figure 1: Long-range dependencies shown as dashed.

Without additional restrictions on the number or placement of the skip edges, exact inference in these models is intractable. For the directed skip-chain model, the tree-width is one more than the maximum number of skip edges passing over a node in the chain. In Sutton and McCallum [2004], loopy belief propagation is used for approximate inference in the skip-chain CRF. In contrast, we introduce an assumption about the structure of the conditional distributions that enables the efficient calculation of posterior marginals.

### 2.3 Mixture-of-parents models

We say that a directed skip-chain model is a mixture-of-parents model if the expression in Equation (1) above can be written in the following special form:

$$p(y_k \mid \mathbf{y}_{\pi_k}, \mathbf{x}) = \sum_{j \in \pi_k} \alpha_{kj} p(y_k \mid y_j, \mathbf{x}), \qquad (2)$$

where $\alpha_{kj} \geq 0$ for each $kj$, and $\sum_{j \in \pi_k} \alpha_{kj} = 1$ are the mixing weights.

We now show that for skip-chain mixture-of-parents models, we can compute the marginal posteriors $p(y_1 \mid \mathbf{x}), \ldots, p(y_n \mid \mathbf{x})$ in an efficient way. In the equations below, all probabilities are conditional on $\mathbf{x}$, so we suppress the $\mathbf{x}$ in our calculations to reduce clutter:

$$p(y_k) = \sum_{y_1,\ldots,y_{k-1}} p(y_k \mid y_1, \ldots, y_{k-1}) \, p(y_1, \ldots, y_{k-1})$$
$$= \sum_{y_1,\ldots,y_{k-1}} \left[ \sum_{j \in \pi_k} \alpha_{kj} p(y_k \mid y_j) \right] p(y_1, \ldots, y_{k-1})$$
$$= \sum_{j \in \pi_k} \sum_{y_1,\ldots,y_{k-1}} \alpha_{kj} p(y_k \mid y_j) \, p(y_1, \ldots, y_{k-1})$$
$$= \sum_{j \in \pi_k} \sum_{y_j} \alpha_{kj} \, p(y_k \mid y_j) \, p(y_j).$$

This calculation shows that if the single-parent conditional probabilities $p(y_k \mid y_j, \mathbf{x})$ are easy to compute, then we can also easily compute the single-node posterior distributions[1]. We can also write the marginal posteriors as

$$p(y_k) = \sum_{j \in \pi_k} \alpha_{kj} p_j(y_k),$$

where $p_j(y_k) = \sum_{y_j} p(y_k \mid y_j) \, p(y_j)$ is the predictive distribution for $y_k$, given the marginal distribution of parent $y_j$. So for a skip-chain mixture-of-parents model, the posterior distribution of a node $y_k$ is a convex combination of the predictive distributions given by each parent separately.

We call this model a skip-chain mixture-of-parents model because Equation (2) defines a probabilistic mixture model. The generative interpretation of such a model is that to generate $y_k$ given the parents $\mathbf{y}_{\pi_k}$, we first randomly choose one of the parent labels according to the multinomial probability distribution with parameters $\alpha_{kj}, j \in \pi_k$. Then, according to the mixture model, only this parent label is relevant to the determination of the label $y_k$. For example, if the randomly chosen parent node is $y_j$, then $y_k$ is drawn according to the conditional probability distribution $p(y_k \mid y_j, \mathbf{x})$.

Conditional distributions with this mixture-of-parents form were also considered in [Pfeffer, 2001], where they were called "separable" distributions. In that work, it is shown that a conditional distribution $p(y_k \mid \mathbf{y}_{\pi_k})$ has the mixture-of-parents form iff we can write the marginal distribution $p(y_k)$ in terms of the marginal distributions of the parents of $y_k$. In general, we would need to know the full joint distribution of the parents to determine the marginal distribution of $y_k$.

### 2.4 Single-parent conditionals

We now complete our description of the mixture-of-parents MEMM by giving the specific form for the individual parent conditional distributions. We use the same maximum entropy model found in the standard

---

[1] Note that the task of finding the posterior mode does not allow the same trick.

MEMM:

$$p_{\lambda,\mu}(y_k \mid y_j, x)$$
$$= \frac{1}{Z_{y_j,\mathbf{x}}} \exp\left(\sum_s \lambda_s f_s(y_j, y_k, \mathbf{x}) + \sum_t \mu_t g_t(y_k, \mathbf{x})\right).$$

Although in theory we could use a different parameter vector $(\lambda, \mu)$ for each edge, in practice we only use a few distinct transition models so that we can pool the data in the parameter estimation phase.

## 3 Learning

We focus on learning the parameters $\lambda$ and $\mu$ of the local transition models $p_{\lambda,\mu}(y_k \mid y_j, \mathbf{x})$, and we assume the mixing weights $\alpha$ are given. In our experiments, we used a uniform mixing distribution.

The standard method for training MEMMs is to maximize the conditional log-likelihood of the data,

$$\mathcal{L}_C(\mathbf{x}, \mathbf{y}) = \log p(\mathbf{y} \mid \mathbf{x}) = \sum_k \log p_{\lambda,\mu}(y_k \mid \mathbf{y}_{\pi_k}, \mathbf{x}),$$

under some regularization of the parameters $\lambda$ and $\mu$. In our experiments, we used ridge regularization, which penalizes the sum of squares of all the weights equally.

The first objective function, $\mathcal{L}_C$, is concave in the parameters $\lambda, \mu$, and therefore easy to optimize. The second objective function, $\mathcal{L}_M$, although not concave, is relatively well-behaved, as noted in Kakade et al. [2002]. In our experiments, we use the L-BFGS [Nocedal and Wright, 1999] method for optimization, which requires us to compute gradients of the objective function.

### 3.1 Gradients

The gradient of the first objective with respect to the parameters $\lambda, \mu$ is given by:

$$\nabla \mathcal{L}_C(\mathbf{x}, \mathbf{y}) = \sum_k \frac{\sum_{j \in \pi_k} \alpha_{kj} \, \nabla p(y_k \mid y_j, \mathbf{x})}{p(y_k \mid \mathbf{y}_{\pi_k}, \mathbf{x})}. \quad (3)$$

For the second objective the gradient is:

$$\nabla \mathcal{L}_M(\mathbf{x}, \mathbf{y}) = \sum_k \frac{\nabla p(y_k \mid \mathbf{x})}{p(y_k \mid \mathbf{x})}. \quad (4)$$

Expanding the gradient of each posterior marginal, we have:

$$\nabla p(y_k \mid \mathbf{x}) = \sum_{j \in \pi_k} \alpha_{kj} \, \nabla p_j(y_k \mid \mathbf{x}). \quad (5)$$

Expanding further, we have:

$$\nabla p_j(y_k \mid \mathbf{x}) = \quad (6)$$
$$\sum_{y'_j} \left[ p(y_k \mid y'_j, \mathbf{x}) \nabla p(y'_j \mid \mathbf{x}) + p(y'_j \mid \mathbf{x}) \nabla p(y_k \mid y'_j, \mathbf{x}) \right].$$

Finally, the gradients of the conditional distributions are

$$\partial_{\mu_t} p(y_k \mid y_j, \mathbf{x}) = \quad (7)$$
$$p(y_k \mid y_j, \mathbf{x}) \left[ g_t(\mathbf{x}, y_k) - \sum_{y'_k} p(y'_k \mid y_j, \mathbf{x}) g_t(\mathbf{x}, y'_k) \right]$$

and

$$\partial_{\lambda_s} p(y_k \mid y_j, \mathbf{x}) = \quad (8)$$
$$p(y_k \mid y_j, \mathbf{x}) \left[ f_s(\mathbf{x}, y_j, y_k) - \sum_{y'_k} p(y'_k \mid y_j, \mathbf{x}) f_s(\mathbf{x}, y_j, y'_k) \right].$$

Note that to calculate the gradient of $\mathcal{L}_M$, we need to compute the marginals $p(y_k \mid \mathbf{x})$, while no inference is required to compute the gradient of $\mathcal{L}_C$. This highlights the difference between the two objectives, as $\mathcal{L}_M$ incorporates the uncertainty in the prediction of previous labels during learning, while $\mathcal{L}_C$ simply uses the correct labels of previous positions. Although it may be advantageous to account for uncertainty in earlier predictions, this makes the gradient calculation at position $k$, as in Equation 5, much more difficult, since we need to incorporate gradients from previous positions. In many natural language tasks, the set of local features that are active (non-zero) at a position is usually small (tens to hundreds). This sparsity of $\nabla p(y'_k \mid y'_j, \mathbf{x})$ allows efficient learning for MEMMs, even with millions of features, since the contributions of each position to the gradient affect only a small number of features. This property no longer holds for the gradient of the $\mathcal{L}_M$ objective, since the gradient contribution of each position $k$ will contain the union of the features active at all of its ancestors' positions. For the $\mathcal{L}_C$ objective, we will have the union of just the parents, not all the ancestors.

### 3.2 Speeding up gradient calculations

The key to efficient gradient computation that exploits sparsity is to reorder the calculations so that only sparse vectors need to be manipulated. If we recursively expand Equations 4, 5, and 6, and regroup the terms, the total gradient can be written as a linear combination of local gradient vectors:

$$\nabla \mathcal{L}_M(\mathbf{x}, \mathbf{y}) = \sum_k \sum_{j \in \pi_k} \sum_{y'_k} \sum_{y'_j} w_{kj}(y'_k, y'_j) \nabla p(y'_k \mid y'_j, \mathbf{x}),$$

where the weights $w_{kj}(y'_k, y'_j)$ depend only on the marginals $p(y_k \mid \mathbf{x})$ and the mixing weights $\alpha_{kj}$. See Appendix A for the derivation of the weights. Thus once we know the weights, the gradient is just the weighted sum of the sparse derivative vectors $\nabla p(y'_k \mid y'_j, \mathbf{x})$. To calculate the weights, we sweep from left to right, computing the appropriate $w_{kj}(y'_k, y'_j)$ recursively. A second left-to-right sweep just adds the sparse gradients with the computed weights.

## 4 The Tasks

We apply the mixture-of-parents MEMM to the CoNLL 2003 English named entity recognition (NER) dataset[2], and the *WebKB* dataset [Craven et al., 1998].

### 4.1 The CoNLL NER task

This NER data set was developed for the shared task of the 2003 Conference on Computational Natural Language Learning (CoNLL). It was one of two NER data sets developed for the task. We used the English language dataset, which comprises Reuters newswire articles annotated with four entity types: location (LOC), person (PER), organization (ORG), and miscellaneous (MISC). The competition scored entity taggers based on their precision and recall at the entity level. Each tagger was ranked based on its overall F1 score, which is the harmonic mean of precision and recall across all entity types. We report this F1 score in our own experiments. We use the standard split of this data into a training set comprising 945 documents and a test set comprising 216 documents.

### 4.2 WebKB

The WebKB dataset contains webpages from four different Computer Science departments: Cornell, Texas, Washington and Wisconsin. Each page is categorized into one of the following five webpage types: course, faculty, student, project, and other. The data set is problematic in that the category other is a diverse mix of many different types of pages. We used the subset of the dataset from Taskar et al. [2002], with the following category distribution: course (237), faculty (148), other (332), research-project (82) and student (542). The number of pages for each school are: Cornell (280), Texas (291), Washington (315) and Wisconsin (454). The number of links for each school are: Cornell (574), Texas (574), Washington (728) and Wisconsin (1614). For each page, we have access to the entire html source, as well as the links to other pages. Our goal is to collectively classify webpages into one of these five categories. In all of our experiments, we learn a model from three schools and test the performance of the learned model on the remaining school.

[2]Available at http://www.cnts.ua.ac.be/conll2003/ner/

## 5 Methods and Results

There are several things one must consider when applying a mixture-of-parents MEMM to data. First, although a MEMM may theoretically have a different parameter vector $(\lambda, \mu)$ for each edge, in practice this gives too many parameters to estimate from the data. The typical approach is to use the same parameter vectors on multiple edges. In terms of the model description above, we limit the number of distinct maximum-entropy conditional probability distributions $p_{\lambda,\mu}(y_k | y_{\pi_k}, \mathbf{x})$ that we must learn. In the NER task, for example, we restrict to two conditional probability models, one that models the transition probability between adjacent words, denoted $p_{\lambda,\mu}(y_k | y_{k-1}, \mathbf{x})$, and another that models the transition probability between nonadjacent words, denoted $p_{\lambda',\mu'}(y_k | y_v, \mathbf{x})$, for $v \leq k - 2$.

Next, one must decide on the edge structure of the graph. That is, for each node in our model, we must have a rule for finding its parent nodes. For sequential data, such as the NER dataset, one obvious parent for a node is the node immediately preceding it. In their skip-chain conditional random field, Sutton and McCallum [2004] put an edge between all pairs of identical capitalized words, in addition to the edges connecting adjacent words.

Once we've specified the parents for every node in the model, we must devise a way to set the mixing weight $\alpha_{kj}$ for the $j$th parent of the $k$th node, for every valid pair $(j, k)$. While one can certainly try to learn a parametric model for the mixing weights, our preliminary results in this direction were not promising. Thus we chose to use uniform mixing weights. That is, we took $\alpha_{kj} = 1/|\pi_k|$, where $|\pi_k|$ denotes the number of parents of node $k$.

Finally, we complete our specification of the maximum-entropy conditional probability distributions by specifying the feature functions $f_s$ and $g_t$. It seems reasonable that the types of features that would be best-suited for a long-distance transition model would be different from the best features for a local model. For example, one reasonable feature for a skip-edge $(y_j, y_k)$ would be whether or not the preceding words $x_{j-1}$ and $x_{k-1}$ are the same. In particular, if the preceding words are equal, this would would make it more likely that the labels agree: $y_j = y_k$. However, this reasoning doesn't apply for local edges: this feature would only be active if the same word occurred twice in a row in the sentence.

Our first objective was to compare mixture-of-parent MEMMs as directly as possible with regular MEMMs. To this end, we took the features on each skip-edge $(y_j, y_k)$ to be the union of the features on the local edges $(y_{j-1}, y_j)$ and $(y_{k-1}, y_k)$. In this first stage, we avoided innovation in the design of skip-edge features since, after all, one could just as well improve plain MEMMs by using better features. However, perhaps we were overly cautious. Although it seems plausible that one could get better results by crafting features particularly well-suited to skip-edges, our initial attempts in this direction did not show any significant improvement. Thus we only report our results using the baseline skip-edge features described above.

### 5.1 The NER Task

For the NER task we used the same feature set used in Sutton and McCallum [2005]. In deciding which skip-edges to include, we first eliminated from consideration all words occurring in more than 100 of the documents. We did this to decrease training time, and because common words are typically easy to label.

After eliminating the most common words from consideration, we followed Sutton and McCallum [2005] and connected the remaining identical capitalized word pairs within each document. However, to keep from having an excessively large training set, if a word occurred more than $r$ times within a *single* document, we only connected it to the $r$ most recent occurrences. The performance of the model seemed relatively insensitive to the value of $r$ between 3 and 10, so we kept it at 5 throughout the experiments.

In Figure 2, we tabulate the performance results of several sequence models on the NER test set. The Viterbi-decoded MEMM is the typical MEMM model. When the mixture-of-parents MEMM has no skip-edges, we get the posterior-decoded MEMM model. These two models perform approximately the same.

If we use the MoP-MEMM model, but train each transition model separately using standard MEMM training, then we get about 0.6% improvement in F1 over the basic MEMM. If we train the models jointly, using the sum of marginal log-likelihoods objective function $\mathcal{L}_M$, then we get an additional 0.4% gain.

The standard CRF model gets an F1 of 90.6%, which even beats the more basic MoP-MEMM model. In principal, one advantage the (undirected) CRF-based models have over the (directed) MEMM-based models is that in the CRF, the label of a given node can be directly influenced by nodes both before and after it. Finally, the skip-chain CRF gets the top performance, beating the jointly trained MoP-MEMM by 0.3%.

| Model | F1% | FP%/FN% | %Improvement |
|---|---|---|---|
| *MEMM* | | | |
| Viterbi | 89.8 | 9.0/11.4 | 2.2/-4.6 |
| Posterior | 89.9 | 9.2/10.9 | 0/0 |
| *Mixture-of-Parents MEMM* | | | |
| Separate | 90.5 | 8.7/10.3 | 5.4/5.5 |
| Joint | 90.9 | 8.5/9.8 | 7.6/10.1 |
| *CRF-Based Models* | | | |
| Linear-Chain | 90.6 | 8.5/10.3 | 7.6/5.5 |
| Skip-Chain | 91.2 | N/A | N/A |

Figure 2: Comparison of several models on the NER task. All models used the feature set described in Sutton and McCallum [2004], which was also the source for the skip-chain CRF result. The F1 column gives the overall entity-wise F1 score, and the FP and FN columns give the entity-wise false positive and false negative rates. The %Improvement column gives the percent reduction in false positive and false negative rates compared to the posterior-decoded MEMM model.

#### 5.1.1 Analysis

It's informative to look in more detail at a particular document in the NER test set. Consider the 15th document, which is an article about a tennis tournament. The tennis player MaliVai Washington is first mentioned, with his full name, in the 17th position of the document. His last name, Washington, shows up 6 more times. The posterior-decoded MEMM correctly labels the first occurrence of Washington as a person with probability 0.99, but 5 of the next 6 occurrences are labeled as locations, though not by a large margin. The Mixture-of-Parents MEMM gets all but one of these 7 occcurrences correct. The very high confidence in the label of the first occurrence of Washington propagates via the skip-edges to later occurrences, tipping things in the right direction.

The improvement in F1 of the MoP-MEMM over the MEMM tells us that the propagation of information via skip edges helps more often than not. Nevertheless, sometime skip-edges lead the model astray. For example, the 22nd document of the test set is about an event in the soccer world. The acronym UEFA, which stands for the Union of European Football Associations, occurs twice in the document. The first time it is correctly identified as an organization by both models, with probability 0.99. The second occurrence of the acronym is in the phrase "the UEFA Cup," which should be a "MISC" entity type. The local edge model indeed predicts that UEFA is a MISC with probability 0.96. However, with uniform mixing, the correct local model in the MoP-MEMM is slightly overpowered by the highly incorrect non-local model prediction, which gives probability 0.99 to the second occurence of UEFA being an ORG.

## 5.2 The WebKB Task

In the WebKB task, the obvious graph structure given by the hyperlinks cannot be taken immediately as our edge model — the problem is that the hyperlink graph may have directed cycles. Our approach to this problem was to first select a random ordering of the nodes. If the $i$'th node and the $j$'th node in our random ordering are connected by an edge (in either direction), with $i < j$, then in our model we put a directed edge from the $i$'th node to the $j$'th node. We use two different conditional probability models for the nodes on these edges. If the original hyperlink is pointing from $i$ to $j$, then we use the "incoming" edge model, and otherwise we use the "outgoing" edge model. In this way, we get a DAG structure with two distinct conditional probability models. Since this is now a MoP-MEMM, we can label the nodes using our standard method.

It's clear that some orderings will give rise to better models than others. To reduce this variability, we find the node marginals resulting from each of 50 random node permutations. We then predict using the average of the 50 marginals. The performance of this average predictor was typically close to the performance of the best of the 50 individual predictors.

We again tried two different approaches to training the conditional probability models. Note that any hyperlink can end up associated with either an incoming or outgoing edge model, depending on the randomly chosen ordering of its nodes. Thus for the "separate training" mode, each hyperlink was added to the training sets of both the incoming and outgoing edge models, and we trained each model using standard MEMM training. For "joint training," we have to fix a random ordering to get a MoP-MEMM model. Again, to reduce variability, we trained on 10 different orderings of the nodes. For this dataset, joint training performed marginally worse than separate training. However, both models essentially matched the performance of the "Link" model of Taskar et al. [2002]. The Link model has a similar graphical structure to our model, but without the mixture-of-parents simplification. Thus to find the labels, they use loopy belief propagation, an approximate inference technique. For comparison, we also trained a maximum-entropy node classifier that ignored the hyperlink information. The performance of this "Node" model, as well as the other models we've discussed, are shown in Figure 3.

| Model | %Error | %Improvement |
|---|---|---|
| Node Model | 16.9 | 0 |
| Link Model | 13.6 | 19.5 |
| MoP-MEMM (*Separate*) | 13.1 | 22.5 |
| MoP-MEMM (*Joint*) | 13.3 | 21.3 |

Figure 3: Comparison of the percent error rates of several models on the WebKB webpage classification task. All models use the feature set described in Taskar et al. [2002], which was also the source for the Link Model result. The %Improvement column gives the percent decrease in the error rate, compared to the node model.

## 6 Discussion

In addition to the works described above (Sutton and McCallum [2004], Finkel et al. [2005]), our work is similar to that of Malouf [2002] and Curran and Clark [2003]. In the latter works, the label of a word is conditioned on the most recent previous instance of that same word in an earlier sentence. To perform inference, they labeled each sentence sequentially, and allowed labels in future sentences to be conditioned on the labels chosen for earlier sentences. The Mixture-of-Parents MEMM seeems to be a conceptual improvement over these methods, since we use the soft labeling (i.e. the posterior distribution) of the preceding label, rather than the predicted label, which doesn't account for the confidence of the labeling.

The skip-chain MEMM has many compelling attributes. First, it is a non-Markovian sequence model for which we can efficiently compute the exact marginal node posteriors. Second, it gives results that exceed ordinary MEMMs, sometimes by a significant margin, without any additional feature engineering. Finally, the model is very modular: We can train several different local and skip-edge models, and interchange them to see which combinations give the best performance. Once a good "separately trained" MoP-MEMM is found, one can use this as a starting point in training the weights of a "jointly trained" MoP-MEMM.

There are also some drawbacks of the skip-chain MEMM, especially in comparison to other non-Markovian models, such as the approaches of Sutton and McCallum [2004] and Finkel et al. [2005]. The most important one is that in skip-chain MEMMs, information only flows from early labels to later labels. Although this may not be a serious problem in the newswire corpora we consider, since earlier mentions of an entity are typically less ambiguous than later mentions, it is certainly less than desirable in general.

One way to address this is by sampling orderings of the nodes and averaging results of inference on several orderings, as we have done for the WebKB hypertext data, showing that a simple mixture of tractable models can capture dependencies in the data as well as an intractable model, which relies on heuristic approximate inference.

## Acknowledgements

We would like to thank the anonymous reviewers and Percy Liang for helpful comments.

## A  Sparse gradient computations

It is easier to work in matrix notation for these derivations. Consider the gradients of the marginals and conditionals with respect to a particular parameter $\theta$ (either $\lambda$ or $\mu$), and for convenience define the following variables:

$$\begin{aligned} v_k(y'_k) &= \partial_\theta p(y'_k \mid \mathbf{x}), \quad \forall k;\ y'_k \in \mathcal{Y}; \quad (9)\\ u_{kj}(y'_j, y'_k) &= \partial_\theta p(y'_k \mid y'_j, \mathbf{x}), \quad \forall k; j \in \pi_k;\ y'_j, y'_k \in \mathcal{Y}. \end{aligned}$$

We stack the vectors $v_1, \ldots, v_n$ into a single vector $\mathbf{v}$ of length $|\mathcal{Y}|n$, where $n$ is number of variables $y_k$ (e.g. the length of the sequence). Similarly, we stack the elements of the $u_{kj}$ matrices into a single vector $\mathbf{u}$ of length $|\mathcal{Y}|^2 n \sum_k |\pi_k|$. By combining Equations (5-6), we can write

$$\mathbf{v} = A\mathbf{v} + B\mathbf{u},$$

for appropriately defined matrices $A$ and $B$. Solving for $\mathbf{v}$, we have $\mathbf{v} = (I - A)^{-1} B\mathbf{u}$. If $\mathbf{v}$ is laid out in blocks corresponding to position $k$, then $A$ is upper triangular with 0's on the diagonal, and thus $I - A$ is easily invertible. The total gradient $\nabla \mathcal{L}_M(\mathbf{x}, \mathbf{y}) = \sum_k v_k(y_k) = \gamma^\top \mathbf{v}$ for an appropriately defined $\gamma$. Hence,

$$\nabla \mathcal{L}_M(\mathbf{x}, \mathbf{y}) = \gamma^\top (I - A)^{-1} B\mathbf{u} = \mathbf{w} \cdot \mathbf{u},$$

where $\mathbf{w} = \gamma^\top (I - A)^{-1} B$ gives the weights $w_{kj}(y'_j, y'_k)$ of local sparse gradients in Section 3.2.